\documentclass{article}
\usepackage{float}
\usepackage{amsmath}
\usepackage{PRIMEarxiv}
\usepackage{makecell}
\usepackage[utf8]{inputenc} 
\usepackage[T1]{fontenc}    
\usepackage{hyperref}       
\usepackage{url}            
\usepackage{booktabs}       
\usepackage{amsfonts}       
\usepackage{nicefrac}    
\usepackage{microtype}      
\usepackage{lipsum}
\usepackage{fancyhdr}       
\usepackage{graphicx}       
\graphicspath{{media/}} 
\usepackage{xcolor}
\usepackage{authblk}
\usepackage{hyperref}
\usepackage{tabularx}  
\usepackage{booktabs}  
\usepackage{geometry}
\usepackage{array} 
\usepackage{ragged2e} 
\usepackage{longtable} 
\usepackage{array} 
\usepackage{ragged2e} 
\usepackage{graphicx}      
\usepackage{subcaption}   
\usepackage{longtable}
\usepackage[utf8]{inputenc}
\usepackage[T1]{fontenc}
\usepackage{natbib}
\begin{document}
\title{Human-Level and Beyond: Benchmarking Large Language Models Against Clinical Pharmacists in Prescription Review}

\author[1,†]{Yan Yang}
\author[2,†]{Mouxiao Bian}
\author[1]{Peiling Li}
\author[1]{Bingjian Wen}
\author[2]{Ruiyao Chen}
\author[2]{Kangkun Mao}
\author[1]{Xiaojun Ye}
\author[2]{Tianbin Li}
\author[2,3]{Pengcheng Chen}
\author[2]{Bing Han}
\author[2,*]{Jie Xu}
\author[1,*]{Kaifeng Qiu}
\author[1,*]{Junyan Wu}
\affil[1]{\textit{
    SUN YAT-SEN MEMORIAL HOSPITAL \\
    Guangdong, China
}}
\affil[2]{\textit{
    Shanghai Artificial Intelligence Laboratory\\
    Shanghai, China
}}
\affil[3]{\textit{
   University of Washington\\
    Washington, USA
}}
\footnotetext[1]{†These authors contributed equally.}
\footnotetext[2]{*Correspondence: 
Junyan Wu(wujunyan@mail.sysu.edu.cn),
Kaifeng Qiu(feng.qk@163.com),
Jie Xu (xujie@pjlab.org.cn)
}

\maketitle
\begin{abstract}
The rapid advancement of large language models (LLMs) has accelerated their integration into clinical decision support, particularly in prescription review. To enable systematic and fine-grained evaluation, we developed RxBench, a comprehensive benchmark that covers common prescription review categories and consolidates 14 frequent types of prescription errors drawn from authoritative pharmacy references. RxBench consists of 1,150 single-choice, 230 multiple-choice, and 879 short-answer items, all reviewed by experienced clinical pharmacists. We benchmarked 18 state-of-the-art LLMs and identified clear stratification of performance across tasks. Notably, Gemini-2.5-pro-preview-05-06, Grok-4-0709, and DeepSeek-R1-0528 consistently formed the first tier, outperforming other models in both accuracy and robustness.  Comparisons with licensed pharmacists indicated that leading LLMs can match or exceed human performance in certain tasks. Furthermore, building on insights from our benchmark evaluation, we performed targeted fine-tuning on a mid-tier model, resulting in a specialized model that rivals leading general-purpose LLMs in performance on short-answer question tasks. The main contribution of RxBench lies in establishing a standardized, error-type–oriented framework that not only reveals the capabilities and limitations of frontier LLMs in prescription review but also provides a foundational resource for building more reliable and specialized clinical tools.
\end{abstract}


\section{Introduction}
Medication errors remain a leading cause of patient harm and healthcare costs. In the United States alone, 7,000-9,000 deaths and over \$40 billion in annual costs are attributed to preventable medication errors\cite{Naseralallah2025}. The prescribing stage is particularly error‐prone. It’s reported that over 75\% of medication errors occur during prescribing or administration\cite{Pais2024}. A study from a pediatric hospital found that clinical pharmacist review detected at least one prescribing error in 81\% of discharge prescriptions\cite{Christiansen2008}. Thus, rigorous review of prescriptions is critical. 

Prescription checking by clinical pharmacists has been shown to improve medication safety and rationality, reduce inappropriate drug use and waste\cite{fan2023prospective}\cite{naseralallah2025role}\cite{skains2025emergency}, and lower rates of  hospital readmissions\cite{costello2025post}\cite{ravn2018effect}. However, the availability of trained pharmacists often lags behind growing clinical demand, especially with the increasing burden of multimorbidity and polypharmacy. In addition, pharmacist-led reviews are time-consuming, highly dependent on individual expertise\cite{cheng2020satisfaction}, and difficult to scale across diverse clinical settings. These constraints contribute to variability in review quality and leave gaps in ensuring medication safety. Collectively, these challenges highlight the urgent need for innovative solutions that can enhance both the efficiency and accuracy of prescription review. 

LLMs have opened new opportunities for prescription review. With their advanced natural language understanding and multi-step reasoning capabilities, LLMs can analyze clinical records and prescriptions while considering the complex relationships among drugs, diseases, and patient-specific characteristics. Recent studies have started to explore their potential in clinical pharmacy tasks. For instance, Huang et al.\cite{Huang2024} compared GPT-4 (ChatGPT) with licensed pharmacists on  pharmacy practice questions, and found that while ChatGPT performed comparably in “medication consultation” (mean scores 8.77 vs. 9.50), it lagged significantly in “prescription review” (5.23 vs. 9.90, p = 0.0089). In another study, a  retrieval-augmented generation (RAG)-based clinical decision support system was developed to detect prescription errors, using 23 complex cases comprising 61 erroneous prescription scenarios. Its co-pilot mode achieved 54.1\% accuracy in the task\cite{Ong2024}. Stansfield et al. \cite{Bull2024}evaluated GPT-4 on the UK Prescribing Safety Assessment and reported an overall accuracy of 79.7\% (153/192) in the prescription review module. Similarly, Li et al. \cite{Li2025}compared eight generative AI systems across key clinical pharmacy tasks, including prescription review, using 48 real-world problems covering ten categories of prescription errors. 

Despite these promising findings, most existing studies treat prescription review merely as a subset of general medical question answering. Evaluations often rely on small-scale, non-specialized datasets, with limited coverage of critical error categories such as off-label use, Inappropriate diluent selection, and skin test requirement labeling. Moreover, current evaluation metrics mostly emphasize binary correctness. There is no standardized benchmark dedicated to systematically assessing LLM performance in prescription review. This absence hinders fair model comparison, obscures the identification of model-specific error patterns.

\begin{figure}
    \centering
    \includegraphics[width=1\linewidth]{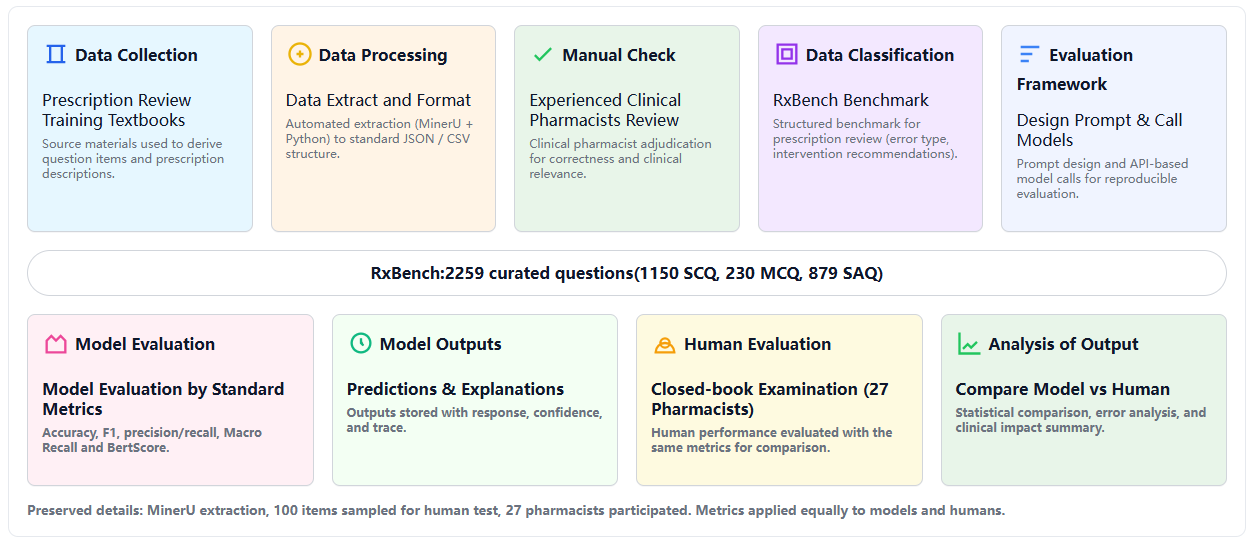}
    \caption{RxBench: End-to-End Pipeline for Prescription Review Benchmarking}
    \label{fig:introduction}
\end{figure}

To address this gap, this work propose RxBench (Figure \ref{fig:introduction}), a comprehensive benchmark specifically designed to evaluate LLMs in prescription review. Our work makes three main contributions: (1) A pharmacist-verified dataset was constructed, comprising diverse prescription cases, including antineoplastic drugs, gynecological endocrine diseases drugs, high-alert medications, high-risk, allergenic drugs, immunosuppressive agents, intravenous medications, off-label drug use, traditional chinese medicine and pharmaceutical compounding;  (2) we conducted baseline evaluations of several state-of-the-art LLMs, providing new insights into their capabilities and limitations; and (3) These  evaluation results directly informed the development of a fine-tuned model for prescription review, which demonstrated enhanced performance in detecting complex medication errors and improves practical applicability. It is  believed that RxBench will serve as a foundational resource for advancing research on LLMs in clinical pharmacy and supporting their safe integration into real-world prescription review workflows.

\section{Methods}
\subsection{Competing Models}
Eighteen representative large language models released from November 2024 to May 2025 were evaluated, spanning the OpenAI, Qwen, Gemini, Claude, and LLaMA families. These models capture a broad spectrum of recent advances (Table \ref{tab:llm_comparison}).
\begin{table}[h!]
\setlength{\tabcolsep}{3pt}
\centering
\caption{Introduction of Large Language Models for Evaluation}
\renewcommand{\arraystretch}{1.2}
\label{tab:llm_comparison}
\begin{tabular}{l l c l c l }
\toprule
\textbf{Model Name} & \textbf{Parameters} & \textbf{Open-Source} & \textbf{Organization} & \textbf{Release Date} & \textbf{Model Type} \\
\midrule
Baichuan2-13B-chat & 13B & Yes & Baichuan AI & Dec 2024 & Text \\
Deepseek-R1-0528 & 671B & Yes & DeepSeek & May 2025 & Text \\
DeepSeek-V3 & 671B & Yes & DeepSeek & Dec 2024 & Text \\
Gemma-3-27B & 27B & Yes & Google & Apr 2025 & Multimodal \\
Llama-4-maverick & 400B & Yes & Meta & Mar 2025 & Multimodal \\
Mistral-small-3.1-24B-instruct & 24B & Yes & Mistral AI & Mar 2025 & Multimodal \\
Qwen2.5-72B-Instruct & 72B & Yes & Alibaba & Nov 2024 & Text \\
Qwen3-14B & 14B & Yes & Alibaba & Apr 2025 & Text \\
Qwen3-32B & 32B & Yes & Alibaba & Apr 2025 & Text \\
Qwen3-235B-A22B & 235B & Yes & Alibaba & Apr 2025 & Text \\

Claude-3-7-sonnet-20250219 & N/A & No & Anthropic & Feb 2025 & Multimodal \\
Claude-sonnet-4-20250514 & N/A & No & Anthropic & May 2025 & Multimodal \\
Gemini-2.0-flash & N/A & No & Google DeepMind & Dec 2024 & Multimodal \\
Gemini-2.5-flash & N/A & No & Google DeepMind & May 2025 & Multimodal \\
GPT-4o-2024-11-20 & 200B & No & OpenAI & Nov 2024 & Multimodal \\
GPT-4.1-2025-04-14 & N/A & No & OpenAI & Apr 2025 & Multimodal \\
Grok-4 & 100--175B & No & xAI & Mar 2025 & Multimodal \\
o4-mini-2025-04-16 & N/A & No & OpenAI & Apr 2025 & Multimodal \\
\bottomrule
\end{tabular}
\end{table}
\subsection{Clinical Pharmacists Test}
To compare model performance with clinical pharmacists across professional levels, we constructed a standardized 100-item test from RxBench, comprising 51 single-choice, 10 multiple-choice, and 39 short-answer questions. Twenty-seven pharmacists from the pharmacy departments of three tertiary hospitals were recruited: 18 pharmacists, 6 senior pharmacists, and 3 principle pharmacists. The assessment was conducted as a two-hour, closed-book, double-blind examination, with all participants completing the test independently. This design ensured rigor and direct comparability between human and model performance.

\subsection{Prescription Error Type}
A standardized assessment framework was developed by reviewing national and international guidelines, clinical practice standards, and regulatory documents, supplemented by practical considerations from clinical care. Fourteen common categories of prescription errors were identified (Table \ref{tab:prescription_errors}).
\begin{longtable}{p{4.5cm} p{10.5cm}}
\caption{Classification and Description of Prescription Errors} \label{tab:prescription_errors} \\
\toprule
\textbf{Error Type} & \textbf{Description} \\
\midrule
\endfirsthead
\toprule
\textbf{Error Type} & \textbf{Description} \\
\midrule
\endhead
\endlastfoot

\textbf{Inappropriate Dosing Regimen} & 
The prescribed dose, frequency, or duration of therapy deviates from established guidelines, potentially compromising therapeutic efficacy, increasing the risk of adverse drug events (ADEs), or leading to other negative clinical outcomes. Examples include exceeding single or daily dose limits, inappropriate dosing intervals, failure to transition from a loading dose to a maintenance dose, and suboptimal or excessive duration of treatment. \\
\midrule

\textbf{Unsuitable Patient Population} & 
The medication is contraindicated or requires caution in the patient's specific demographic or physiological state, such as in pediatric, geriatric, pregnant, or lactating populations, or in patients with significant hepatic or renal impairment. \\
\midrule

\textbf{Unwarranted Indication} & 
The therapeutic indication for the prescribed medication is inconsistent with the patient's clinical diagnosis, treatment goals, or disease state. \\
\midrule

\textbf{Inappropriate Dosage Form or Route of Administration} & 
The selected dosage form or route of administration is not aligned with the product's labeling, clinical practice guidelines, or the patient's condition, which may lead to diminished efficacy or adverse reactions. Examples include prescribing solid oral dosage forms to patients with dysphagia or selecting an intravenous route for a drug intended for intramuscular administration. \\
\midrule

\textbf{Incompatibility or Clinically Significant Interaction} & 
Co-administration of two or more drugs results in physical or chemical incompatibility (e.g., precipitation, degradation) or a clinically significant drug-drug interaction that alters the pharmacokinetics or pharmacodynamics of one or more agents, leading to reduced efficacy or increased toxicity. \\
\midrule

\textbf{Inappropriate Solvent or Vehicle} & 
The selection or volume of the solvent or vehicle used for drug reconstitution or dilution does not adhere to the manufacturer's instructions or established pharmaceutical standards, posing a risk of altered drug stability, precipitation, or adverse reactions. \\
\midrule

\textbf{Therapeutic Duplication} & 
The concurrent, unjustifiable prescribing of two or more medications with identical or overlapping pharmacological mechanisms of action. For instance, prescribing both acetaminophen and a combination cold product containing acetaminophen. \\
\midrule

\textbf{Inappropriate Timing of Administration} & 
The scheduling of drug administration is suboptimal, potentially affecting bioavailability, efficacy, or tolerability, or fails to adhere to critical timing requirements. Examples include administering a gastrointestinal irritant on an empty stomach or failing to administer surgical antibiotic prophylaxis within the recommended 0.5-2 hour pre-incisional window. \\
\midrule

\textbf{Contraindicated Drug Use} & 
The patient's clinical condition is a contraindication to the drug as listed in the prescribing information. For example, prescribing aspirin to a patient with an active peptic ulcer. \\
\midrule

\textbf{Omission of Required Allergy Test Documentation} & 
For medications necessitating a preliminary allergy or skin test, the medical record lacks documentation confirming that the test was performed and its outcome. \\
\midrule

\textbf{Prescribing to a Patient with a Known Allergy} & 
A medication is prescribed to a patient with a documented history of allergy to the drug itself or to a structurally related compound with a known risk of cross-reactivity. \\
\midrule

\textbf{Suboptimal Drug Selection} & 
While the drug is indicated for the patient's diagnosis, a more appropriate agent exists based on patient-specific factors (e.g., comorbidities, concomitant medications, genetic profile), and the current choice may result in inferior outcomes or higher risk. \\
\midrule

\textbf{Prescription Non-conformance}& 
The prescription fails to meet formal requirements as stipulated by regulatory standards (e.g., "Management Specification for Hospital Prescription Review"). This includes, but is not limited to: 1. Omissions, non-standard formatting, or illegibility in prescription components; 2.Non-compliant or inconsistent physician signature/seal; 3.Absence of a pharmacist's appropriateness review; 4. Omission of age in days/months for neonatal/infant prescriptions; 5.Failure to issue separate prescriptions for different drug categories; 6.Use of non-proprietary or non-standard drug names; 7.Ambiguous or non-standard notation for dose, strength, quantity, or units; 8.Use of vague instructions such as "as directed" or "for personal use."\\
\midrule

\textbf{Off-label Use} & 
The use of a drug outside the scope of its marketing authorization as approved by the national regulatory authority. This encompasses use for an unapproved indication, patient population, dosage, route of administration, frequency, or duration of therapy. \\
\bottomrule

\end{longtable}

\subsection{LoRA Fine-tuning Method}
To adapt the pre-trained language model for the specialized domain of prescription review, we employ the Low-Rank Adaptation (LoRA) method for parameter-efficient fine-tuning. LoRA introduces low-rank decomposed adapter modules while keeping the pre-trained weights frozen, enabling targeted adjustment of model behavior.

\subsubsection{Model Configuration}
We implement LoRA fine-tuning within the Megatron-LM framework with the following specifications:
\begin{itemize}
    \item \textbf{Rank Parameters}: LoRA rank is set to 32 with a scaling factor $\alpha$ of 64, ensuring a balance between parameter efficiency and representational capacity.
    \item \textbf{Target Modules}: LoRA adapters are injected into all linear layers (all-linear), including key transformation matrices in attention mechanisms and feed-forward networks.
    \item \textbf{Training Strategy}: We employ a gradient accumulation strategy with global batch size of 16 and micro batch size of 8, combined with sequence parallelism and Flash Attention backend to optimize memory usage.
\end{itemize}

\subsubsection{Training Parameters}
The model is trained for 10 epochs using a constant learning rate of $10^{-4}$ with 5\% warmup proportion. To prevent overfitting, the learning rate linearly decays to a minimum of $10^{-5}$. We adopt a full recomputation strategy with uniform gradient checkpointing per layer, effectively balancing computational overhead and memory consumption.

\subsection{Statistical Analysis}
All statistical analyses were conducted using Python. To assess overall performance differences among multiple models, the Friedman test was employed as a non-parametric alternative to repeated-measures ANOVA. Pairwise comparisons of classification outcomes, particularly in error-type analyses visualized through heatmaps, were evaluated using McNemar’s test. A two-tailed $P$ value of 0.05 was considered the threshold for statistical significance. Unless otherwise specified, all reported results include exact $P$ values when significant differences were observed.For the pharmacist groups, results were aggregated and reported as mean scores within each professional rank (pharmacist, senior pharmacist, and principle pharmacist) to ensure comparability with model outputs.
\section{Dataset and Experiment}
\subsection{Data Source and Expert Review}
All test and training items were derived from authoritative clinical pharmacist training textbooks. The scope of prescription types was broad and clinically representative, covering medications used in cardiovascular diseases, gastrointestinal disorders, otorhinolaryngologic conditions, infectious diseases, chronic diseases in older adults, pregnancy and lactation, pain management, pediatrics, renal disorders, respiratory diseases, endocrine and metabolic diseases, as well as neurological and psychiatric conditions.

The test dataset consisted of both single-choice questions, multiple-choice questions and short-answer questions (Figure \ref{fig:introduction}), whereas the training dataset included only short-answer questions, totaling 2,547 items. All items, regardless of disease system or question format, were mapped to a unified set of error categories as defined in Table \ref{tab:prescription_errors} and subsequently underwent manual review.

To ensure data quality, a review committee was convened, consisting of three senior pharmacists with over five years of experience in prescription review and one principal pharmacist with more than ten years of experience. Inclusion criteria required completeness of both questions and reference answers, clinical representativeness, educational relevance, and reflection of common or high-risk prescription errors. Exclusion criteria included missing information, outdated regimens inconsistent with current guidelines, duplicate or highly similar entries, and items limited to rote memorization without clinical relevance.

Each item was independently reviewed and cross-checked against reference answers, followed by group discussion to resolve discrepancies. This process ensured that the dataset accurately captured the reasoning and judgment required in real-world prescription review, thereby guaranteeing its scientific validity and consistency.

In addition, to facilitate automated evaluation, clinical pharmacists extracted key answer points from the reference solutions to generate predefined scoring rubrics.

\subsection{Evaluation Tasks Design}
To objectively assess model performance, a zero-shot evaluation paradigm was adopted. For multiple-choice questions, prompts were restricted to elicit only the final answer to ensure accuracy in subsequent analysis. For short-answer questions, where established paradigms are lacking, prompts explicitly defined error categories and their descriptions, while enforcing a standardized output format. Detailed prompt designs for each item type are provided in Table \ref{tab:prompt_definitions}.
\begin{longtable}{p{4.5cm} p{10.5cm}}

\caption{Prompts of Different Tasks} \label{tab:prompt_definitions} \\

\toprule
\textbf{Category} & \textbf{Prompt} \\
\midrule
\endfirsthead

\toprule
\textbf{Category} & \textbf{Prompt} \\
\midrule
\endhead

\midrule
\endlastfoot

\textbf{Single-choice question} & 
You are a clinical pharmacist with extensive expertise. Your task is to answer the following single-choice question based on your knowledge of evidence-based pharmacy. Output only the letter of the most appropriate answer; do not include any other content. Example output: \texttt{A} \\
\midrule

\textbf{Multiple-choice question} &
You are a clinical pharmacist with extensive expertise. Your task is to answer the following multiple-choice question based on your knowledge of evidence-based pharmacy. Output all letters of the correct answers; do not include any other content. Example output: \texttt{ABC} \\
\midrule

\textbf{Short-answer question} & 
\textbf{Background:} \par
You are a clinical pharmacist with extensive expertise. Your task is to conduct a systematic review of electronic prescriptions based on evidence-based pharmacy principles to identify, prevent, and resolve potential or actual drug-related problems, ensuring patient medication is safe, effective, and rational. \par\medskip
\textbf{Core Task:} \par
Please analyze the provided \texttt{[Prescription Information]}, strictly adhering to the \texttt{[Problem Type Definitions and Explanations]} to identify and classify any issues. Finally, generate a professional and rigorous prescription review report according to the specified \texttt{[Output Format]}. \par\medskip

\textbf{[Problem Type Definitions and Explanations]} \par
See Table\textbf{~\ref{tab:prescription_errors} }for error types and their definitions. \par\medskip

\textbf{[Output Format]} \textbf{Strictly adhere to this format for your response. Do not output any extraneous content or explanations.} \par\medskip
--If the review identifies problems, use this format: \par
\textbf{Problem Type:} [Select the most accurate classification(s) from the list above. Separate multiple types with a semicolon. Example: Inappropriate Dosing Regimen; Off-label Use] \par
\textbf{Intervention Suggestion:} [Briefly describe the specific issues in the prescription and the proposed interventions.] \par\medskip
--If the review finds no issues, use this format: \par
\textbf{Prescription without Discrepancies} \\

\end{longtable}
\subsection{Evaluation Framework and Procedure}
For each task, we constructed a complete input consisting of the problem and its prompt: single-choice questions included the stem and options, while short-answer questions included the patient’s baseline information and prescription details. These inputs, combined with task-specific prompts, were submitted to each model via API calls to obtain the outputs, which were then stored in structured JSON files for subsequent analysis. To ensure stable outputs that accurately reflected the models’ intrinsic capabilities, the temperature parameter was fixed at 0, while all other parameters remained at their default settings.For the pharmacist cohort, we collected responses from 18 pharmacists, 6 senior pharmacists, and 3 principle pharmacists. A total of 25 valid responses were obtained for the single- and multiple-choice tasks, and 27 valid responses were collected for the short-answer tasks.

\subsection{Data Post-processing}
Because the prompts enforced a standardized response format, model outputs required minimal post-processing, except for occasional issues such as redundant line breaks. Clinical pharmacists’ responses were collected using Excel spreadsheets; we performed light normalization by correcting mixed Chinese–English punctuation and concatenating content into structured formats. No additional processing was applied beyond these steps.

\subsection{Evaluation Metrics}
Different evaluation methods were adopted for different types of questions, as detailed in Table\ref{tab:evaluation metrics} .
\begin{table}[htbp]
\centering
\caption{Evaluation Metrics for Different Task Types}
\label{tab:evaluation metrics}
\renewcommand{\arraystretch}{1.2}
\begin{tabular}{p{4cm} p{9cm}}
\hline
\textbf{Task Type} & \textbf{Evaluation Metric(s)} \\
\hline
Single-choice questions & Accuracy (based on TP, TN, FP, FN) \\
Multiple-choice questions & F1 score (harmonic mean of Precision and Recall)\&Accuracy (based on TP, TN, FP, FN)\\
Short-answer questions & F1 score (error-type classification); Macro Recall (coverage of intervention points),BERTScore (semantic similarity of intervention); final weighted score\\
\hline
\end{tabular}
\end{table}
\subsubsection{Single-choice questions} 
Performance was evaluated using accuracy (Equation \ref{Accuracy} ), calculated from true positives (TP), true negatives (TN), false positives (FP), and false negatives (FN).  
\begin{equation}
\label{Accuracy}
\text{Accuracy} = \frac{\text{TP} + \text{TN}}{\text{TP} + \text{TN} + \text{FP} + \text{FN}}
\end{equation}
\subsubsection{Multiple-choice questions} 
Performance was measured by the F1 score (Equation \ref{F1_score}), combining precision and recall and accuracy (Equation \ref{Accuracy} ).
\begin{equation}
\label{F1_score}
F_1 = 2 \times \frac{\text{Precision} \times \text{Recall}}{\text{Precision} + \text{Recall}}
\end{equation}

Where:\\
- Precision represents the proportion of samples predicted as positive that are actually positive(Equation \ref{precision}):
 \begin{equation} 
 \label{precision}
  \text{Precision} = \frac{TP}{TP + FP}
  \end{equation}
- Recall represents the proportion of actual positive samples that are correctly predicted as positive(Equation \ref{recall}):
\begin{equation}
 \label{recall}
  \text{Recall} = \frac{TP}{TP + FN}
  \end{equation}

In the above formulas, $TP$ stands for True Positives, $FP$ stands for False Positives, and $FN$ stands for False Negatives.
\subsubsection{Short-answer questions} 
A multidimensional evaluation framework was employed. Error-type classification was assessed using the F1 score (Equation \ref{F1_score}), while the quality of intervention recommendations was evaluated by Macro Recall (calculating  the average recall rate of all answer points within each question by Python) and BERTScore (semantic similarity with references). A weighted integration of these metrics yielded the final score (Equation \ref{total_cal}). Specially, Incorrect error-type classification resulted in a total score of zero.
\begin{equation}
 \label{total_cal}
Total = 0.4 \times Total_{\text{Error Type}} + 0.6 \times Total_{\text{Intervention}}
 \end{equation}
\section{Results}
\subsection{Composition of RxBench}
\begin{figure}
    \centering
    \includegraphics[width=1\linewidth]{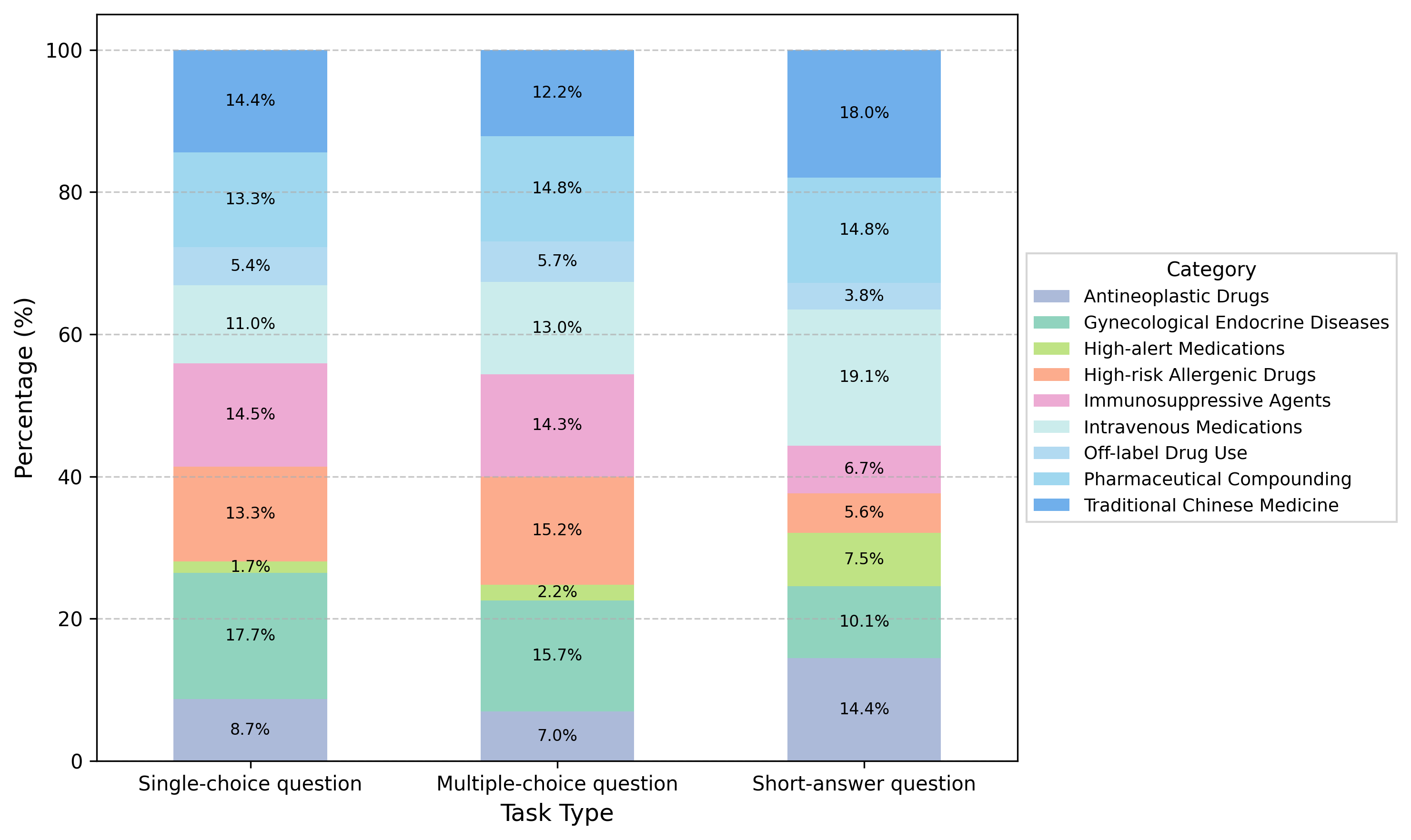}
    \caption{Percentage Distribution of Categories by Task Type}
    \label{fig1}
\end{figure}

Due to the lack of evaluation criteria related to prescription review, we selected relevant content from the prescription review training textbooks for clinical pharmacists and developed RxBench, which is also the first of its kind. RxBench includes two types of items: multiple-choice questions and short-answer questions, covering nine common prescription review types. The multiple-choice section contains 1,150 single-choice and 230 multiple-choice items, each with a stem and five options (A–E). The short-answer section comprises 879 items integrating patient demographics and prescription details. An overview is shown in Figure \ref{fig1}.

\subsection{Large Language Models Comparison}
\subsubsection{Single-Choice Question Task}
\begin{figure}
    \centering
    \includegraphics[width=1\linewidth]{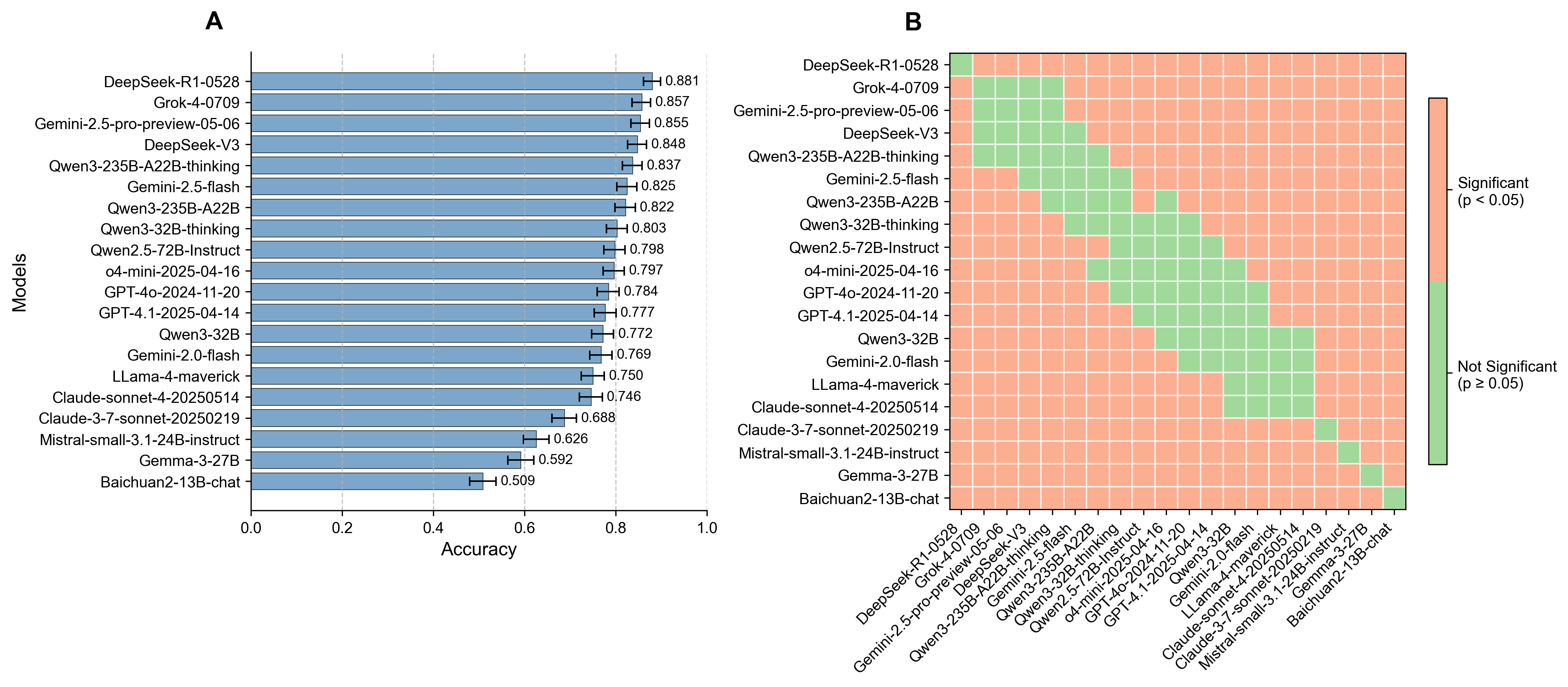}
    \caption{Comparison of LLMs in Single-Choice Question Task}
    \label{fig:single-choice_QA}
\end{figure}

In the single-choice prescription review task, the 18 large language models exhibited substantial variability in accuracy (range: 0.509–0.881, $P < 0.001$; Figure\ref{fig:single-choice_QA}A). DeepSeek-R1-0528 achieved the highest accuracy (0.881), significantly surpassing all other models (McNemar test, all $P < 0.05$). The next tier comprised Grok-4-0709 (0.857) and Gemini-2.5-pro-preview-05-06 (0.855), followed by DeepSeek-V3 (0.848), Qwen3-235B-A22B-thinking (0.837), and Gemini-2.5-flash (0.825). A mid-range cluster—including Qwen3-32B, Gemini-2.0-flash, LLaMA-4-maverick, Claude, and GPT-4 series—achieved accuracies of 0.75–0.80. Performance declined markedly for Mistral-small-3.1-24B-instruct (0.626) and Gemma-3-27B (0.592), with Baichuan2-13B-chat (0.509) performing the worst. Within the Qwen series, “thinking” variants consistently outperformed their standard counterparts. Pairwise McNemar tests showed non-significant differences among the top models (e.g., Grok-4-0709 vs. Gemini-2.5-pro-preview-05-06; Figure\ref{fig:single-choice_QA}B), indicating shared error tendencies at the highest performance level.

\subsubsection{Multiple-Choice Question Task}
\begin{figure}
    \centering
    \includegraphics[width=1\linewidth]{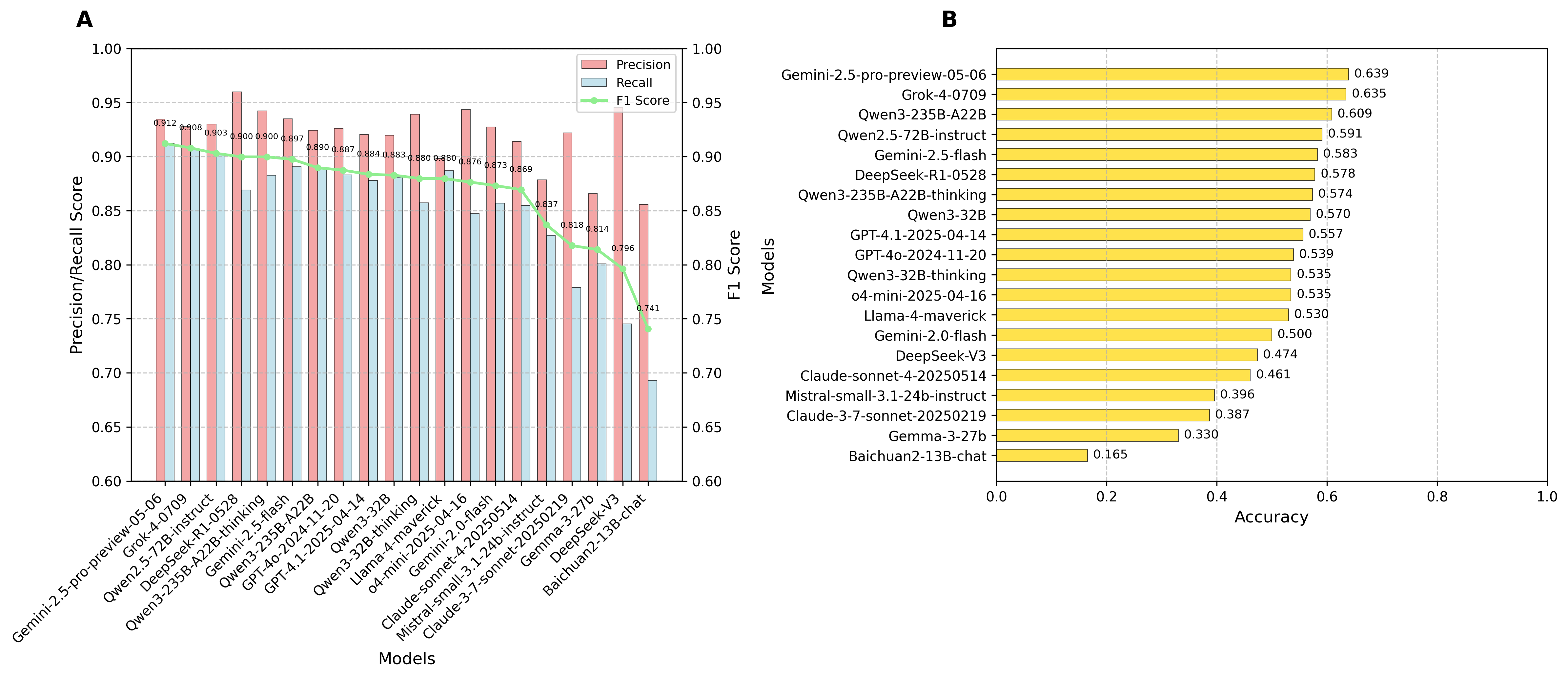}
    \caption{Comparison of LLMs in Multiple-Choice Question Task}
    \label{fig:multiple-choice_QA}
\end{figure}

In the multiple-choice prescription review task, models displayed pronounced divergence in precision, recall, and F1 performance ($P < 0.001$,Figure \ref{fig:multiple-choice_QA}A). The strongest overall performers were Gemini-2.5-pro-preview-05-06, Grok-4-0709, and Qwen2.5-72B-instruct. Gemini-2.5-pro-preview-05-06 led across all three metrics, demonstrating balanced accuracy and coverage. Grok-4-0709 achieved a comparable F1 score, reflecting a favorable balance between recall and precision. Qwen2.5-72B-instruct., although marginally lower, maintained consistently strong performance, underscoring its robustness.

Distinct optimization strategies were evident. Models such as DeepSeek-R1-0528 and DeepSeek-V3 emphasized precision at the expense of recall, reflecting a conservative error-avoidance approach. In contrast, Grok-4-0709 and Qwen3-235B-A22B-thinking prioritized recall, capturing a broader range of correct options while tolerating modest precision losses. These divergent strategies highlight context-dependent trade-offs in model design.

Accuracy results (Figure\ref{fig:multiple-choice_QA}B) reinforced these trends: Gemini-2.5-pro-preview-05-06 achieved the highest accuracy (0.639), closely followed by Grok-4-0709 (0.635). By contrast, Baichuan2-13B-chat performed poorly (0.165), indicating profound limitations in multi-answer contexts. Several models in the Qwen series achieved mid-range accuracy but demonstrated stable F1 scores, suggesting moderate generalizability and transfer potential.In multiple-choice question tasks, thinking models did not seem to absolutely outperform non-thinking models as expected.
\subsubsection{Short-Answer Question Task}
\begin{figure}
    \centering
    \includegraphics[width=1\linewidth]{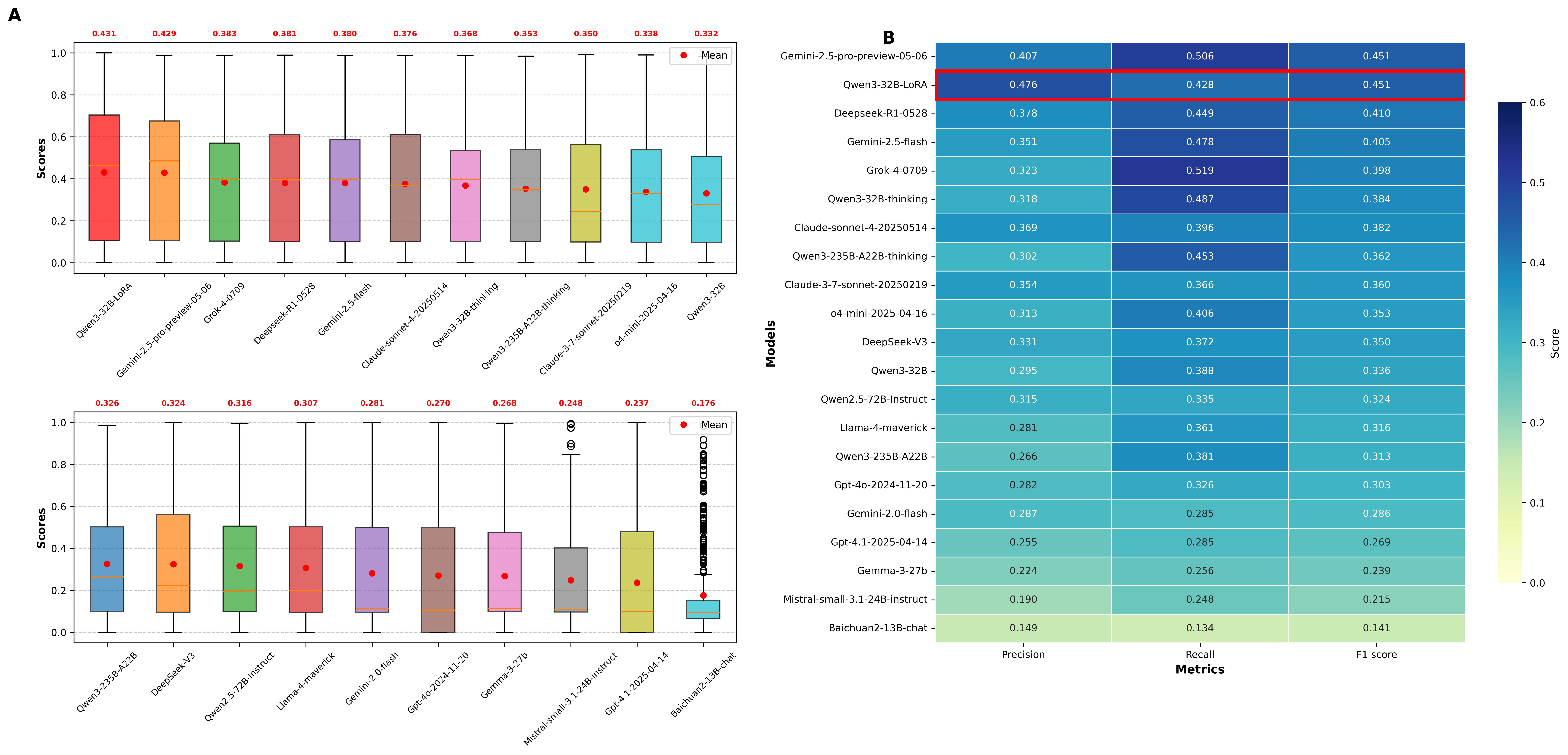}
    \caption{Comparison of LLMs in Short-Answer Question Task}
    \label{fig:Short-Answer_Question}
\end{figure}

In the short-answer prescription review task, models exhibited substantial heterogeneity in overall scoring and error-type recognition ($P < 0.001$). As shown in the boxplots (Figure\ref{fig:Short-Answer_Question}A), Gemini-2.5-pro-preview-05-06, Grok-4-0709, and DeepSeek-R1-0528 consistently achieved higher mean scores with narrower distributions, indicating greater stability and reliability. Conversely, Baichuan2-13B-Chat and Mistral-small-3.1-24B-instruct displayed lower mean scores and greater variability, reflecting limited robustness and inconsistent task adaptation.  

Heatmap analyses of F1 scores across error categories (Figure\ref{fig:Short-Answer_Question}B) further confirmed these trends. Gemini-2.5-pro-preview-05-06 achieved high values across precision, recall, and F1 score (with its F1 score being the highest among all models), reflecting balanced strengths in identifying error types while minimizing over-prediction. Grok-4-0709 demonstrated the highest recall among all models, alongside a relatively high F1 score, though its precision was moderate — indicating heightened sensitivity to detecting errors but reduced precision in predicting errors. DeepSeek-R1-0528 maintained balanced and consistently strong performance across precision, recall, and F1 metrics. In contrast, Baichuan2-13B-Chat showed poor results across all three indicators, underscoring its limited applicability in specialized prescription review contexts.
\subsection{Comparison Between Large Language Models and Pharmacists}

\subsubsection{Single-Choice Question Task}
\begin{figure}
    \centering
    \includegraphics[width=1\linewidth]{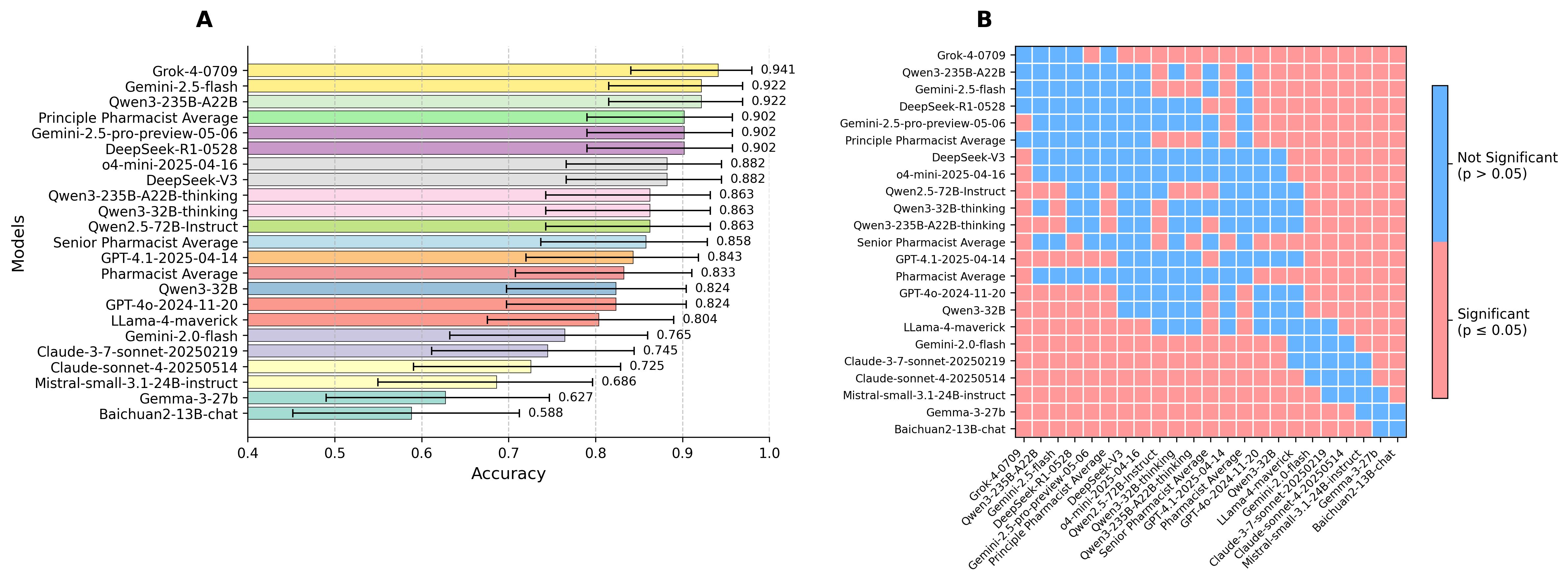}
    \caption{Comparison Between Large Language Models and Pharmacists in Single-Choice Question Task}
    \label{fig:Single-choice_VS}
\end{figure}

In the single-choice prescription review task, clear performance stratification was observed across language models and pharmacist groups. As shown in Figure~\ref{fig:Single-choice_VS}A, pharmacist accuracy ranged from 0.833 (pharmacists) to 0.902 (principle pharmacists). Several advanced models, including Grok-4-0709 (0.941), Qwen3-235B-A22B (0.922), and Gemini-2.5-flash (0.922), exceeded the highest pharmacist benchmark, suggesting superior adaptability to this task. A number of other frontier models (e.g., Gemini-2.5-pro-preview-05-06, DeepSeek-R1-0528, o4-mini-2025-04-16) also achieved comparable accuracy to senior pharmacists (0.858–0.902). In contrast, mid-tier and smaller models, such as Baichuan2-13B-chat (0.588), Gemma-3-27b (0.627), and Mistral-small-3.1-24b-instruct (0.686), performed substantially worse, indicating limited robustness for complex clinical decision-making.

Figure~\ref{fig:Single-choice_VS}B presents the McNemar’s test results for pairwise comparisons. Compared with pharmacists, Grok-4-0709 demonstrated statistically significant superiority over all three ranks (principle pharmacists, senior pharmacists, and pharmacists, $p \leq 0.05$). Qwen3-235B-A22B and Gemini-2.5-flash also significantly outperformed pharmacists and senior pharmacists, and achieved parity with principle pharmacists, showing no significant difference in that comparison ($p > 0.05$). Models in the intermediate range (e.g., GPT-4.1-2025-04-14, Qwen3-32B) generally aligned with pharmacist performance, often showing no significant difference from senior pharmacists or pharmacists, but still trailing behind principle pharmacists. In contrast, low-performing models (e.g., Baichuan2-13B-chat, Gemma-3-27b) were significantly worse than all pharmacist groups. Importantly, no significant difference was observed among pharmacist groups themselves, confirming consistent performance across levels of professional seniority.
\subsubsection{Multiple-Choice Question Task}
\begin{figure}
    \centering
    \includegraphics[width=1\linewidth]{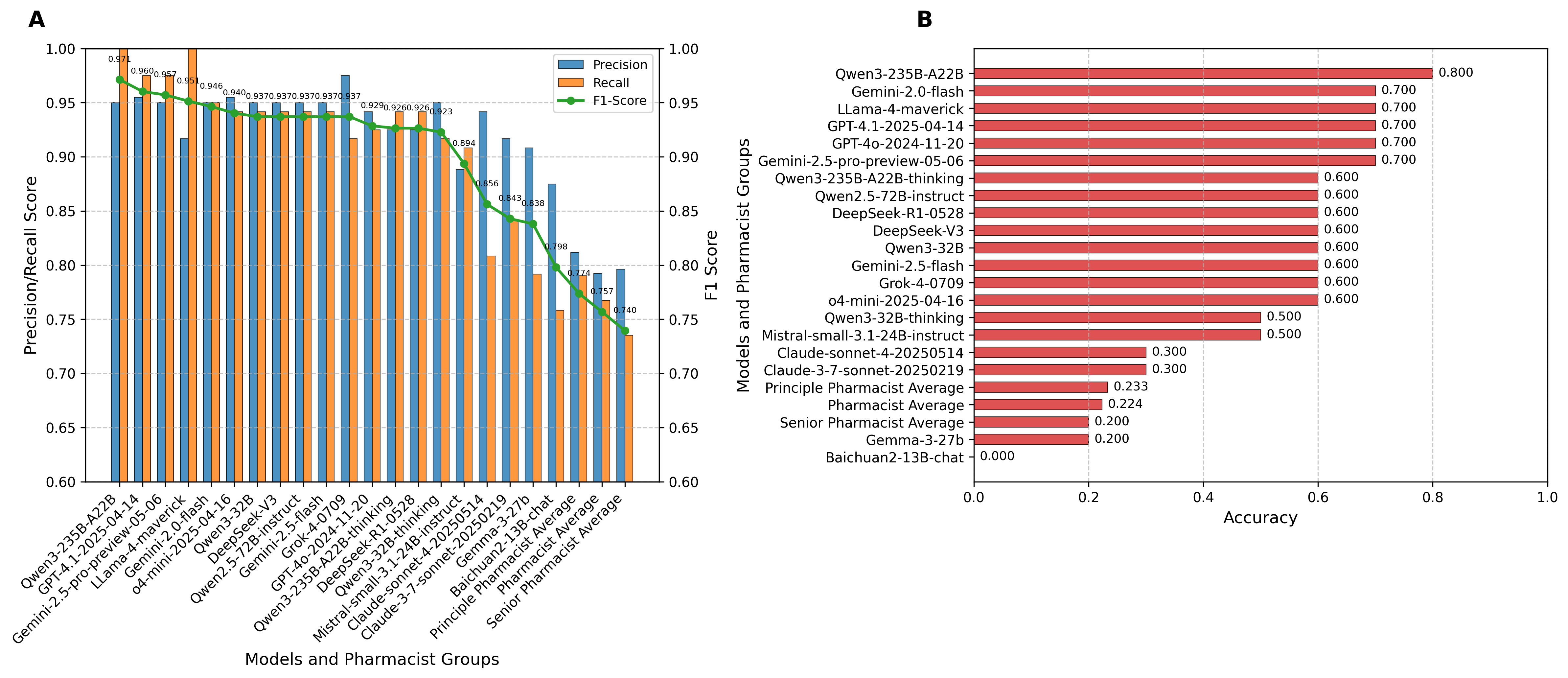}
    \caption{Comparison Between Large Language Models and Pharmacists in Multiple-Choice Question Task}
    \label{fig:Multiple-Choice_VS}
\end{figure}

In the multiple-choice prescription review task, large performance gaps emerged between models and pharmacist groups. As shown in Figure~\ref{fig:Multiple-Choice_VS}A, top-performing models such as Qwen3-235B-A22B, Gemini-2.0-flash, and GPT-4.1-2025-04-14 achieved F1 scores above 0.94, with most leading models maintaining scores above 0.92. In contrast, pharmacist groups performed substantially worse, with F1 scores of 0.757 (principle pharmacists), 0.740 (senior pharmacists), and 0.774 (pharmacists), all well below the mid-tier model range (approximately 0.85–0.90). Notably, the expected gradient by professional rank was absent, as pharmacists slightly outperformed senior pharmacists, suggesting that greater clinical experience did not translate into superior task performance.

Accuracy analysis (Figure~\ref{fig:Multiple-Choice_VS}B) reinforced this pattern. Qwen3-235B-A22B achieved the highest accuracy (0.800), followed by a cluster of high-performing models such as Gemini-2.0-flash, LLaMA-4-maverick, and GPT-4.1-2025-04-14, all at 0.700. Pharmacist groups, however, were positioned near the bottom: 0.233 for principle pharmacists, 0.224 for pharmacists, and 0.200 for senior pharmacists. These values were close to or even below some lower-tier models (e.g., Claude-3.7-sonnet, Mistral-small-3.1), underscoring the striking magnitude of the human–model disparity.

Together, these findings highlight two key insights: (1) state-of-the-art large language models exhibit strong robustness in handling knowledge-intensive, multi-solution tasks, consistently outperforming pharmacists across both F1 and accuracy; and (2) performance stratification by professional seniority was not evident, indicating that pharmacist experience alone was insufficient to achieve higher performance in this evaluation.
\subsubsection{Short-Answer Question Task}
\begin{figure}
    \centering
    \includegraphics[width=1\linewidth]{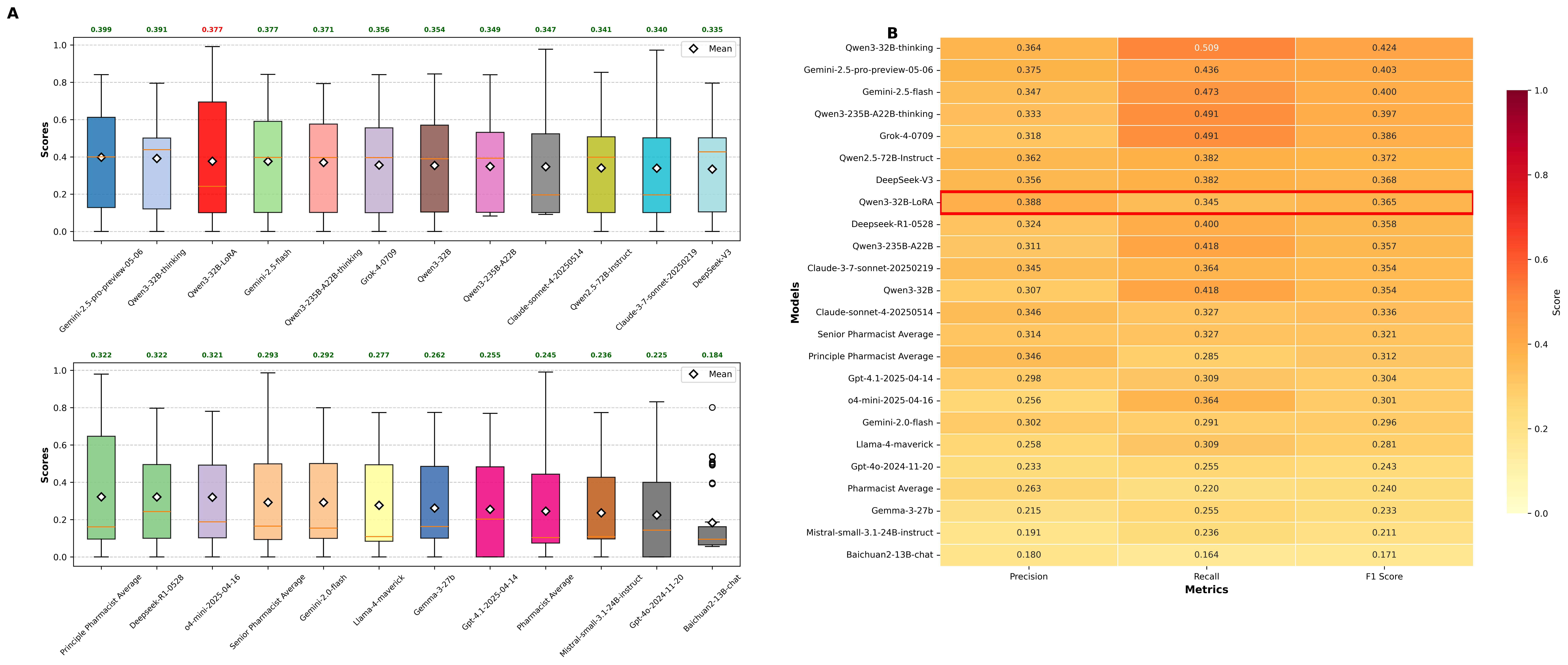}
    \caption{Comparison Between Large Language Models and Pharmacists in Short-Answer Question Task}
    \label{fig:Short-Answer_VS}
\end{figure}

In the short-answer prescription review task, boxplot comparisons (Figure~\ref{fig:Short-Answer_VS}A) revealed substantial variation across models and pharmacist groups. The best-performing models included Gemini-2.5-pro-preview-05-06 (mean score 0.431), DeepSeek-R1-0528 (0.391), and Grok-4-0709 (0.390), which consistently achieved higher averages with relatively stable distributions. By contrast, models such as Baichuan2-13B-chat (0.222) and GPT-4o-2024-11-20 (0.230) scored lowest, showing both weaker central tendencies and higher variability. Pharmacist groups—principle pharmacist (0.322), senior pharmacist (0.316), and pharmacist (0.245)—were positioned in the mid-to-lower range, outperforming some underperforming models but falling behind the leading systems, suggesting that clinical experience provided limited advantage in this task.

The heatmap of precision, recall, and F1 score (Figure~\ref{fig:Short-Answer_VS}B) further illustrated this disparity. Gemini-2.5-pro-preview-05-06 achieved the highest F1 (0.487), followed by Grok-4-0709 (0.435), Qwen3-32B-thinking (0.412), and DeepSeek-R1-0528 (0.410). In comparison, pharmacists scored notably lower, with F1 values of 0.312 (principle pharmacist), 0.321 (senior pharmacist), and 0.240 (pharmacist). The weakest performers, including Baichuan2-13B-chat (0.220) and GPT-4o-2024-11-20 (0.248), showed deficiencies across all three metrics.

Overall, these findings indicate that only the most advanced models demonstrated clear superiority over pharmacists in short-answer tasks, whereas lower- and mid-tier models often underperformed or showed comparable results. This suggests that while cutting-edge systems can enhance error recognition and response accuracy, pharmacist expertise remains competitive relative to many existing models.
\subsection{Performance Improvement After LoRA Fine-Tuning in Short-Answer Question Task}

We selected the moderately performing Qwen3-32B as the base model for fine-tuning using the training dataset. After LoRA adaptation, Qwen3-32B-LoRA achieved a score nearly 30\% higher than the non–fine-tuned Qwen3-32B and 17\% higher than Qwen3-32B equipped only with the fine-tuned thinking mode (Figure~\ref{fig:Short-Answer_Question}A). Notably, the LoRA-enhanced model also surpassed several leading large-parameter models, ranking first overall in the short-answer evaluation.
Qwen3-32B-LoRA demonstrated a substantial improvement in recognition accuracy of error-type, with its F1 score rising to a level second only to Gemini-2.5-pro-preview-05-06 (Figure~\ref{fig:Short-Answer_Question}B).

Similarly, in the comparison with clinical pharmacists, Qwen3-32B-LoRA outperformed the non–fine-tuned Qwen3-32B in both total score and F1 score of error-type(Figure \ref{fig:Short-Answer_VS}). These results collectively indicate that the model retains considerable room for further improvement in prescription-review tasks, underscoring the importance of continued task-specific optimization in future work.
\section{\textbf{Discussion}}

Our study provides a comprehensive evaluation of LLMs and pharmacists in prescription review  involving single-choice, multiple-choice, and short-answer question tasks. The results consistently demonstrate that state-of-the-art LLMs not only reach but often exceed the accuracy and stability of human experts, revealing both opportunities and challenges in integrating these systems into clinical pharmacy practice. 

\subsection{Performance Superiority of LLMs Over Pharmacists}
Across all tasks, advanced LLMs such as Grok-4 and Gemini-2.5-pro-preview-05-06  significantly outperformed pharmacists in accuracy, F1 scores, and overall stability. Particularly in the multiple-choice task, the gap was striking, with models achieving F1 scores exceeding 0.92 while pharmacists remained below 0.78, and accuracy differences reaching nearly fourfold. These findings align with prior research indicating that LLMs are capable of integrating vast knowledge bases and reasoning across complex contexts, often surpassing domain specialists in specific constrained tasks \citep{kung2023performance, gilson2023does,singhal2023large}. A plausible explanation for this performance gap lies in the specialized nature of clinical pharmacy practice. Unlike LLMs, which apply a uniform reasoning process across all domains, clinical pharmacists typically undergo training focused on specific therapeutic areas. While this cultivates deep expertise within their specialty, it may lead to less familiarity or reduced confidence when addressing prescription issues outside their immediate domain, thereby constraining their performance on a broad and diverse question set.

\subsection{Clinical Experience Versus Computational Reasoning}
One unexpected observation was the inverted effect among pharmacists: senior pharmacists did not consistently outperform their junior counterparts. This result suggests that prescription review ability is not determined solely by years of experience, but may also be shaped by task type, evaluation criteria, and cognitive or environmental factors. For instance, different question formats may elicit distinct reasoning strategies. When confronted with complex prescription scenarios—particularly those involving off-label drug use—senior or clinical pharmacists tend to engage in more profound and specialized reasoning, conducting comprehensive risk assessments and evidence-based analyses. However, this cautious and thorough approach may lead to more conservative or hesitant response strategies in a standardized multiple-choice testing format, potentially impacting their scores. By contrast, less experienced pharmacists may rely more strictly on guideline adherence or test-oriented reasoning, which can be advantageous in standardized assessments. LLMs further underscore this divergence, as they excel in structured question-answering tasks by leveraging pattern recognition and probabilistic reasoning, unaffected by clinical experience or contextual nuance. In addition, potential influences such as cognitive bias, workload, or contextual distractions cannot be ruled out. These findings highlight the complexity of evaluating pharmacists’ prescription review competence. Future assessment designs should therefore adopt more diverse and context-oriented tasks, integrating both standardized tests and real-world scenarios, to more comprehensively capture professional performance and provide a basis for targeted training and decision-support system development.

\subsection{Model Strategies and Error Profiles}
Our analyses also revealed divergent optimization strategies among LLMs. Some models (e.g., Claude-sonnet series) prioritized precision at the expense of recall, adopting a conservative stance that minimized false positives, while others (e.g., Grok-4, Qwen3-235B-A22B) emphasized broader recall, achieving higher sensitivity but reduced specificity. This reflects trade-offs similar to those reported in diagnostic AI systems, where performance varies according to the weighting of false-positive versus false-negative risks \citep{rajpurkar2022ai, miotto2018deep}. Notably, heatmap analyses confirmed that models surpassed pharmacists in error-type recognition, albeit with universally modest F1 scores, suggesting an area where both human and machine performance require improvement.

\subsection{\textbf{Effectiveness of Domain-Specific Fine-Tuning}}
Our study demonstrates that domain-specific fine-tuning is a highly effective strategy for overcoming the limitations of general-purpose LLMs in clinical reasoning tasks. Leading LLMs showed significant weaknesses in short-answer prescription review, exposing a core bottleneck in complex clinical reasoning. To address this gap, we conducted supervised fine-tuning of Qwen3-32B, a mid-performing model in our initial assessment, using a rigorously curated dataset of 2,457 high-quality prescription review cases. These cases, sourced from multiple medical centers across China, cover a wide range of clinical departments and have been thoroughly validated by pharmaceutical experts. Each case includes detailed reasoning logic, forming a "textbook-grade" training resource. The fine-tuned model demonstrated remarkable improvement. Notably, it not only significantly surpassed its base version but also outperformed most leading general-purpose models that lacked specialized optimization. This finding strongly suggests that for knowledge-intensive and safety-critical tasks like prescription review, the construction of high-quality, logically-annotated training data may be as important as simply scaling up base model parameters.Furthermore, the suboptimal performance of general-purpose models underscores their lack of targeted training in this specific domain, highlighting the need for future development of specialized LLMs dedicated to prescription review.

\subsection{Implications for Clinical Practice}
The consistent outperformance of pharmacists by LLMs raises important implications for the future role of AI in pharmacy practice. On the one hand, LLMs can serve as powerful decision-support tools, offering enhanced error detection and reducing cognitive load for human pharmacists. Prior work has shown that AI-assisted prescription review can lower dispensing errors and improve workflow efficiency \citep{shoaran2020ai, grzybowski2020artificial}. On the other hand, caution is warranted. LLMs remain prone to hallucination, lack explainability, and may underperform in real-world contexts requiring integration of patient-specific factors such as comorbidities or social determinants of health \citep{wang2024opportunities}. Therefore, careful implementation is essential to ensure that AI systems complement, rather than replace, pharmacists’ professional judgment, thereby supporting clinical decision-making and safeguarding patient safety.

\subsection{Limitations and Future Directions}
Several limitations should be acknowledged. First, the study is constrained by its reliance on cases sourced from eight authoritative clinical pharmacist training textbooks. While this multi-center collection from 29 clinical centers ensures a high standard of accuracy and credibility, the standardized nature of textbook cases may not fully capture the nuances and heterogeneity of real-world clinical practice.the tasks employed here were knowledge- and text-based, which may not fully capture the complexity of real-world clinical decision-making. Second, the evaluation was limited to current benchmark models and may not reflect future iterations with improved reasoning or multimodal capabilities \citep{thirunavukarasu2023large}. Finally, while statistical superiority of LLMs was established, their long-term reliability, interpretability, and integration into pharmacy workflows require further investigation in prospective clinical trials.

Future work therefore prioritize validation using real-world data to assess model performance in authentic scenarios. Moreover, research should focus on hybrid human–AI systems that combine the contextual awareness and ethical reasoning of pharmacists with the computational breadth and consistency of LLMs. This complementarity could optimize prescription safety, enhance efficiency, and preserve the pharmacist’s central role in patient care.

\subsection{Conclusion}
In summary, this study demonstrates that cutting-edge LLMs outperform pharmacists across a range of prescription review tasks, with particularly pronounced advantages in complex multiple-choice and error-recognition scenarios. These findings highlight the transformative potential of LLMs as decision-support tools in clinical pharmacy, while underscoring the importance of careful integration, interpretability, and collaborative practice models to ensure safe and effective implementation.

\section{\textbf{Data Availability Statement}
}
The dataset supporting the findings of this study has been made publicly available through the MedBench repository. It can be accessed at \href{https://medbench.opencompass.org.cn/home}{https://medbench.opencompass.org.cn/home}.
\bibliography{references.bib} 

@article{Naseralallah2025,
  author = {Naseralallah, L. and Koraysh, S. and Alasmar, M. and Aboujabal, B.},
  title = {The role of pharmacists in mitigating medication errors in the perioperative setting: a systematic review},
  journal = {Systematic Reviews},
  year = {2025},
  volume = {14},
  number = {1},
  pages = {12},
  doi = {10.1186/s13643-024-02710-1},
  url = {https://doi.org/10.1186/s13643-024-02710-1}
}

@article{Pais2024,
  author = {Pais, C. and Liu, J. and Voigt, R. and Gupta, V. and Wade, E. and Bayati, M.},
  title = {Large language models for preventing medication direction errors in online pharmacies},
  journal = {Nature Medicine},
  year = {2024},
  volume = {30},
  number = {6},
  pages = {1574-1582},
  doi = {10.1038/s41591-024-02933-8},
  url = {https://doi.org/10.1038/s41591-024-02933-8}
}

@article{Christiansen2008,
  author = {Christiansen, S. R. and Morgan, J. A. and Hilmas, E. and Shepardson, A.},
  title = {Impact of a prescription review program on the accuracy and safety of discharge prescriptions in a pediatric hospital setting},
  journal = {J Pediatr Pharmacol Ther},
  year = {2008},
  volume = {13},
  number = {4},
  pages = {226-232},
  doi = {10.5863/1551-6776-13.4.226},
  url = {https://doi.org/10.5863/1551-6776-13.4.226}
}

@article{Huang2024,
  author = {Huang, X. and Estau, D. and Liu, X. and Yu, Y. and Qin, J. and Li, Z.},
  title = {Evaluating the performance of ChatGPT in clinical pharmacy: A comparative study of ChatGPT and clinical pharmacists},
  journal = {Br J Clin Pharmacol},
  year = {2024},
  volume = {90},
  number = {1},
  pages = {232-238},
  doi = {10.1111/bcp.15896},
  url = {https://doi.org/10.1111/bcp.15896}
}

@article{Ong2024,
  author = {Ong, J. C. L. and Jin, L. and Elangovan, K. and Lim, G. Y. S. and Lim, D. Y. Z. and Sng, G. G. R. and Ke, Y. and Tung, J. Y. M. and Zhong, R. J. and Koh, C. M. Y.},
  title = {Development and testing of a novel large language model-based clinical decision support systems for medication safety in 12 clinical specialties},
  journal = {arXiv Preprint},
  year = {2024},
  eprint = {2402.01741},
  archivePrefix = {arXiv},
  url = {https://arxiv.org/abs/2402.01741}
}

@article{Bull2024,
  author = {Bull, D. and Okaygoun, D.},
  title = {Evaluating the Performance of ChatGPT in the Prescribing Safety Assessment: Implications for Artificial Intelligence-Assisted Prescribing},
  journal = {Cureus},
  year = {2024},
  volume = {16},
  number = {11},
  pages = {e73003},
  doi = {10.7759/cureus.73003},
  url = {https://doi.org/10.7759/cureus.73003}
}

@article{Li2025,
  author = {Li, L. and Du, P. and Huang, X. and Zhao, H. and Ni, M. and Yan, M. and Wang, A.},
  title = {Comparative Analysis of Generative Artificial Intelligence Systems in Solving Clinical Pharmacy Problems: Mixed Methods Study},
  journal = {JMIR Med Inform},
  year = {2025},
  volume = {13},
  pages = {e76128},
  doi = {10.2196/76128},
  url = {https://doi.org/10.2196/76128}
}

@article{gilson2023does,
  author = {Gilson, A. and Safranek, C. W. and Huang, T. and Socrates, V. and Chi, L. and Taylor, R. A. and Chartash, D.},
  title = {Does ChatGPT have the potential to support clinical decision-making?},
  journal = {Journal of the American Medical Informatics Association},
  year = {2023},
  volume = {30},
  number = {9},
  pages = {1724–1732},
  doi = {10.1093/jamia/ocad083},
  url = {https://doi.org/10.1093/jamia/ocad083}
}

@article{kung2023performance,
  author = {Kung, T. H. and Cheatham, M. and Medenilla, A. and Sillos, C. and De Leon, L. and Elepaño, C. and others and Tseng, V.},
  title = {Performance of ChatGPT on USMLE: Potential for AI-assisted medical education using large language models},
  journal = {PLOS Digital Health},
  year = {2023},
  volume = {2},
  number = {2},
  pages = {e0000198},
  doi = {10.1371/journal.pdig.0000198},
  url = {https://doi.org/10.1371/journal.pdig.0000198}
}

@article{rajpurkar2022ai,
  author = {Rajpurkar, P. and Chen, E. and Banerjee, O. and Topol, E. J.},
  title = {AI in health and medicine},
  journal = {Nature Medicine},
  year = {2022},
  volume = {28},
  number = {1},
  pages = {31–38},
  doi = {10.1038/s41591-021-01614-0},
  url = {https://doi.org/10.1038/s41591-021-01614-0}
}

@article{miotto2018deep,
  author = {Miotto, R. and Wang, F. and Wang, S. and Jiang, X. and Dudley, J. T.},
  title = {Deep learning for healthcare: Review, opportunities and challenges},
  journal = {Briefings in Bioinformatics},
  year = {2018},
  volume = {19},
  number = {6},
  pages = {1236–1246},
  doi = {10.1093/bib/bbx044},
  url = {https://doi.org/10.1093/bib/bbx044}
}

@article{shoaran2020ai,
  author = {Shoaran, M. and Makin, J. G. and others and Carmena, J. M.},
  title = {AI in pharmacovigilance and prescription safety: A review},
  journal = {Drug Safety},
  year = {2020},
  volume = {43},
  number = {5},
  pages = {421–432},
  doi = {10.1007/s40264-020-00932-9},
  url = {https://doi.org/10.1007/s40264-020-00932-9}
}

@article{grzybowski2020artificial,
  author = {Grzybowski, A. and Brona, P. and Lim, G.},
  title = {Artificial intelligence for medical decision support in ophthalmology},
  journal = {Translational Vision Science \& Technology},
  year = {2020},
  volume = {9},
  number = {2},
  pages = {14–14},
  doi = {10.1167/tvst.9.2.14},
  url = {https://doi.org/10.1167/tvst.9.2.14}
}

@article{thirunavukarasu2023large,
  author = {Thirunavukarasu, A. J. and Ting, D. S. and Elangovan, K. and Gutierrez, L. and Keane, P. A.},
  title = {Large language models in medicine},
  journal = {Nature Medicine},
  year = {2023},
  volume = {29},
  number = {8},
  pages = {1930–1940},
  doi = {10.1038/s41591-023-02448-8},
  url = {https://doi.org/10.1038/s41591-023-02448-8
        
        
        
        
        
        
        
        
        }      
}

@article{wang2024opportunities,
  author = {Wang, F. and Wang, Y. and Rastegar-Mojarad, M. and Liu, S.},
  title = {Opportunities and challenges of LLMs in healthcare},
  journal = {npj Digital Medicine},
  year = {2024},
  volume = {7},
  pages = {55},
  doi = {10.1038/s41746-024-01055-w},
  url = {https://doi.org/10.1038/s41746-024-01055-w}        
}

@article{singhal2023large,
  title={Large language models encode clinical knowledge},
  author={Singhal, Karan and Azizi, Shekoofeh and Tu, Tao and Mahdavi, S Sara and Wei, Jason and Chung, Hyung Won and Scales, Nathan and Tanwani, Ajay and Cole-Lewis, Heather and Pfohl, Stephen and others},
  journal={Nature},
  volume={620},
  number={7972},
  pages={172--180},
  year={2023},
  publisher={Nature Publishing Group}
}

@article{fan2023prospective,
  title={Prospective prescription review system correlated with more rational PPI medication use, better clinical outcomes and reduced PPI costs: experience from a retrospective cohort study},
  author={Fan, Xiucong and Chen, Danxia and Bao, Siwei and Dong, Xiaohui and Fang, Fang and Bai, Rong and Zhang, Yuyi and Zhang, Xiaogang and Tang, Weijun and Ma, Yabin and others},
  journal={BMC Health Services Research},
  volume={23},
  number={1},
  pages={1014},
  year={2023},
  publisher={Springer}
}

@article{naseralallah2025role,
  title={The role of pharmacists in mitigating medication errors in the perioperative setting: a systematic review},
  author={Naseralallah, Lina and Koraysh, Somaya and Alasmar, May and Aboujabal, Bodoor},
  journal={Systematic Reviews},
  volume={14},
  number={1},
  pages={12},
  year={2025},
  publisher={Springer}
}

@article{skains2025emergency,
  title={Emergency department programs to support medication safety in older adults: a systematic review and meta-analysis},
  author={Skains, Rachel M and Hayes, Jane M and Selman, Katherine and Zhang, Yue and Thatphet, Phraewa and Toda, Kazuki and Hayes, Bryan D and Tayes, Carla and Casey, Martin F and Moreton, Elizabeth and others},
  journal={JAMA network open},
  volume={8},
  number={3},
  pages={e250814--e250814},
  year={2025},
  publisher={American Medical Association}
}

@article{costello2025post,
  title={A post-discharge pharmacist clinic to reduce hospital readmissions: a retrospective cohort study},
  author={Costello, Jaclyn and Barras, Michael and Snoswell, Centaine L and Foot, Holly},
  journal={International Journal of Clinical Pharmacy},
  pages={1--9},
  year={2025},
  publisher={Springer}
}

@article{ravn2018effect,
  title={Effect of an in-hospital multifaceted clinical pharmacist intervention on the risk of readmission: a randomized clinical trial},
  author={Ravn-Nielsen, Lene Vestergaard and Duckert, Marie-Louise and Lund, Mia Lolk and Henriksen, Jolene Pilegaard and Nielsen, Michelle Lyndgaard and Eriksen, Christina Skovsende and Buck, Thomas Croft and Potteg{\aa}rd, Anton and Hansen, Morten Rix and Hallas, Jesper},
  journal={JAMA internal medicine},
  volume={178},
  number={3},
  pages={375--382},
  year={2018},
  publisher={American Medical Association}
}

@article{cheng2020satisfaction,
  title={Satisfaction and needs of pharmacists in prescription-checking training: a cross-sectional survey},
  author={Cheng, Wei and Wang, Chen and Ma, Jing and Ji, Wen and Yang, Xiangli and Wu, Bei and Hou, Ruigang},
  journal={Journal of International Medical Research},
  volume={48},
  number={11},
  pages={0300060520965810},
  year={2020},
  publisher={SAGE Publications Sage UK: London, England}
}
\bibliographystyle{plainnat}

\end{document}